\documentclass[11pt,a4paper,table]{article}
\usepackage[hyperref]{acl2019}
\usepackage{times}
\usepackage{latexsym}

\usepackage[T5,T2A,T1]{fontenc}
\usepackage[utf8]{inputenc}

\usepackage{url}
\usepackage{calc}
\usepackage{xcolor}
\usepackage{graphicx}
\usepackage{multirow}
\usepackage{amsthm}
\usepackage{amssymb}
\usepackage{amsmath}
\usepackage{amsfonts}
\usepackage{color}
\usepackage{mdwlist}
\usepackage{microtype}
\usepackage{tabularx}
\usepackage{booktabs}
\usepackage{subcaption}
\usepackage{covington}
\usepackage{makecell}
\usepackage{arydshln}

\definecolor{Gray}{gray}{0.9}
\definecolor{LightGray}{gray}{0.88}
\definecolor{LighterGray}{gray}{0.92}
\usepackage{enumitem,amssymb}
\usepackage[russian,english]{babel}
\usepackage{CJKutf8}

\usepackage{float}

\newlist{todolist}{itemize}{2}
\setlist[todolist]{label=$\square$}
\usepackage{pifont}

\aclfinalcopy 

\newcommand\matt[1]{\textcolor{blue}{matt: \textbf{#1}}}

\title{Training Neural Response Selection for Task-Oriented Dialogue Systems}

\author{
 Matthew Henderson,
 Ivan Vuli\'{c},
 Daniela Gerz,
 I{\~{n}}igo Casanueva,
 Pawe{\l} Budzianowski, \\
 {\bf
 Sam Coope,
 Georgios Spithourakis,
 Tsung-Hsien Wen,
 Nikola Mrk{\v{s}}i\'c,
 \textmd{and} Pei-Hao Su
 } \vspace{2mm} \\
 PolyAI Limited \\
 London, United Kingdom \\
 \texttt{\small matt@poly-ai.com \hspace{2mm} ivan@poly-ai.com}
}

\begin{document}
\maketitle
\begin{abstract}
Despite their popularity in the chatbot literature, retrieval-based models have had modest impact on task-oriented dialogue systems, with the main obstacle to their application being the low-data regime of most task-oriented dialogue tasks. Inspired by the recent success of pretraining in language modelling, we propose an effective method for deploying \emph{response selection} in task-oriented dialogue. To train response selection models for task-oriented dialogue tasks, we propose a novel method which: \textbf{1)} pretrains the response selection model on large general-domain conversational corpora; and then \textbf{2)} fine-tunes the pretrained model for the target dialogue domain, relying only on the small in-domain dataset to capture the nuances of the given dialogue domain. Our evaluation on six diverse application domains, ranging from e-commerce to banking, demonstrates the effectiveness of the proposed training method. 
\end{abstract}

\section{Introduction}
\label{s:intro}
\begin{figure}[!t]
\centering
\includegraphics[width=1.0\linewidth]{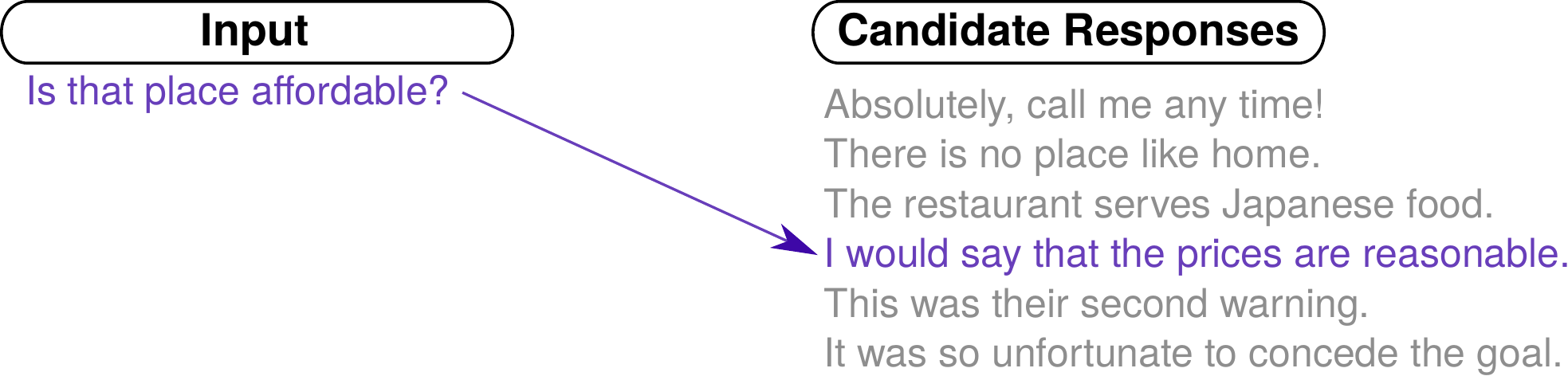}
\vspace{-4.5mm}
\caption{The conversational response selection task: given the input sentence, the goal is to identify the relevant response from a large collection of candidates.}
\vspace{-12.5mm}
\label{fig:task}
\end{figure}

Retrieval-based dialogue systems conduct conversations by selecting the most appropriate system \text{response} given the dialogue history and the input user {utterance} (i.e., the full dialogue {context}). A typical retrieval-based approach to dialogue encodes the input and a large set of responses in a joint semantic space. When framed as an ad-hoc retrieval task \cite{Deerwester:1990jasis,Ji:2014arxiv,Kannan:2016kdd,Henderson:2017arxiv}, the system treats each input utterance as a \textit{query} and retrieves the most relevant response from a large response collection by computing semantic similarity between the query representation and the encoding of each response in the collection. This task is referred to as \textit{response selection} \cite{Wang:2013emnlp,AlRfou:2016arxiv,Yang:2018repl,Du:2018scai,Chaudhuri:2018conll,Weston:2018ws}, as illustrated in Figure~\ref{fig:task}.

Formulating dialogue as a response selection task stands in contrast with other data-driven dialogue modeling paradigms such as modular and end-to-end task-based dialogue systems \cite{young:10b,Wen:17,Liu:2017eacl,Li:2017ijcnlp,Bordes:2017iclr}. Unlike standard task-based systems, response selection does not rely on explicit task-tailored semantics in the form of domain ontologies, which are hand-crafted for each task by domain experts \cite{Henderson:14a,Henderson:14b,Mrksic:15}. Response selection also differs from chatbot-style systems which generate new responses by generalising over training data, their main deficiency being the tendency towards generating universal but irrelevant responses such as \textit{``I don't know''} or \textit{``Thanks''} \cite{vinyals:15,Li:2016naacl,Serban:2016aaai,Song:2018ijcai}. Therefore, response selection removes the need to engineer structured domain ontologies, and to solve the difficult task of general language generation. Furthermore, it is also much easier to constrain or combine the output of response selection models. This design also bypasses the construction of dedicated decision-making policy modules. 


Although conceptually attractive, retrieval-based dialogue systems still suffer from \textit{data scarcity}, as deployment to a new domain requires a sufficiently large in-domain dataset for training the response selection model. Procuring such data is expensive and labour-intensive, with annotated datasets for task-based dialogue still few and far between, as well as limited in size.\footnote{For instance, the recently published MultiWOZ dataset \cite{Budzianowski:2018emnlp} comprises a total of 115,424 dialogue turns scattered over 7 target domains. It is several times larger than other standard task-based dialogue datasets such as DSTC2 \cite{Henderson:14b} with 23,354 turns, Frames \cite{ElAsri:2017sigdial} with 19,986 turns, or M2M \cite{Shah:2018naacl} with 14,796 turns. To illustrate the difference in magnitude, the Reddit corpus used in this work for response selection pretraining comprises 727M dialogue turns.}


Recent work on language modelling (LM) pretraining \cite{Peters:2018naacl,Howard:2018acl} has shown that task-specific architectures are not necessary in a number of NLP tasks. The best results have been achieved by LM pretraining on large unannotated corpora, followed by supervised fine-tuning on the task at hand \cite{Devlin:2018arxiv}. Given the compelling benefits of large-scale pretraining, our work poses a revamped question for response selection:  can we \textit{pretrain} a general {response selection model} and then \emph{adapt} it to a variety of different dialogue domains? 


To tackle this problem, we propose a two-step training procedure which: \textbf{1)} pretrains a response selection model on large conversational corpora (such as Reddit); and then \textbf{2)} fine-tunes the pretrained model for the target dialogue domain. Throughout the evaluation, we aim to provide answers to the following two questions:

\begin{enumerate}
    \item \textbf{(Q1)} \emph{How to pretrain?} Which encoder structure can best model the Reddit data? 
    \item  \textbf{(Q2)} \textit{How to fine-tune?} Which method can efficiently adapt the pretrained model to a spectrum of target dialogue domains?   
\end{enumerate}



Regarding the first question, the results support findings from prior work \cite{Cer:2018arxiv,Yang:2018repl}: the best scores are reported with simple transformer-style architectures \cite{Vaswani:2017nips} for input-response encodings. Most importantly, our results suggest that pretraining plus fine-tuning for response selection is useful across six different target domains. 

As for the second question, the most effective training schemes are lightweight: the model is pretrained only once on the large Reddit training corpus, and the target task adaptation does not require expensive retraining on Reddit. We also show that the proposed two-step response selection training regime is more effective than directly applying off-the-shelf state-of-the-art sentence encoders \cite{Cer:2018arxiv,Devlin:2018arxiv}. 

We hope that this paper will inform future development of response-based task-oriented dialogue. Training and test datasets, described in more detail by \newcite{Henderson:2019arxiv}, are available at: {
    \ttfamily \small \href{https://github.com/PolyAI-LDN/conversational-datasets}{github.com/{\allowbreak}PolyAI-LDN/conversational-datasets}}.

\section{Methodology}
\label{s:methodology}
\paragraph{Why Pretrain and Fine-Tune?}
By simplifying the conversational learning task to a response selection task, we can relate target domain tasks to general-domain conversational data such as Reddit~\cite{AlRfou:2016arxiv}. This also means that parameters of response selection models in target domains with scarce training resources can be initialised by a general-domain pretrained model. 

The proposed two-step approach, described in \S\ref{ss:pretraining} and \S\ref{ss:fine-tuning}, can be seen as a ``lightweight'' task adaptation strategy: the expensive Reddit model pretraining is run only once (i.e., training time is typically measured \textit{in days}), and the model is then fine-tuned on $N$ target tasks (i.e., fine-tuning time is \textit{in minutes}). The alternatives are ``heavyweight'' data mixing strategies. First, in-domain and Reddit data can be fused into a single training set: besides expensive retraining for each task, the disbalance between in-domain and Reddit data sizes effectively erases the target task signal. An improved data mixing strategy keeps the identities of the origin datasets (Reddit vs. target) as features in training. While this now retains the target signal, our preliminary experiments indicated that the results again fall short of the proposed lightweight fine-tuning strategies. In addition, this strategy still relies on expensive Reddit retraining for each task.

\subsection{Step 1: Response Selection Pretraining}
\label{ss:pretraining}
\paragraph{Reddit Data.}
Our pretraining method is based on the large Reddit dataset compiled and made publicly available recently by \newcite{Henderson:2019arxiv}. This dataset is suitable for response selection pretraining due to multiple reasons as discussed by \newcite{AlRfou:2016arxiv}. First, the dataset offers organic conversational structure and it is large at the same time: all Reddit data from January 2015 to December 2018, available as a BigQuery dataset, span almost 3.7B comments. After preprocessing the dataset to remove both uninformative and long comments\footnote{We retain only sentences containing more than 8 and less than 128 word tokens.} and pairing all comments with their responses, we obtain more than 727M comment-response pairs which are used for model pretraining. This Reddit dataset is substantially larger than the previous Reddit dataset of \newcite{AlRfou:2016arxiv}, which spans around 2.1B comments and 133M conversational threads, and is not publicly available. Second, Reddit is extremely diverse topically \cite{Schrading:2015emnlp,AlRfou:2016arxiv}: there are more  than  300,000 sub-forums (i.e., subreddits) covering diverse topics of discussion. Finally, compared to message-length-restricted Twitter conversations \cite{Ritter:2010naacl}, Reddit conversations tend to be more natural. In summary, all these favourable properties hold promise to support a large spectrum of diverse conversational domains.

\begin{figure}[!t]
\centering
\includegraphics[width=1.0\linewidth]{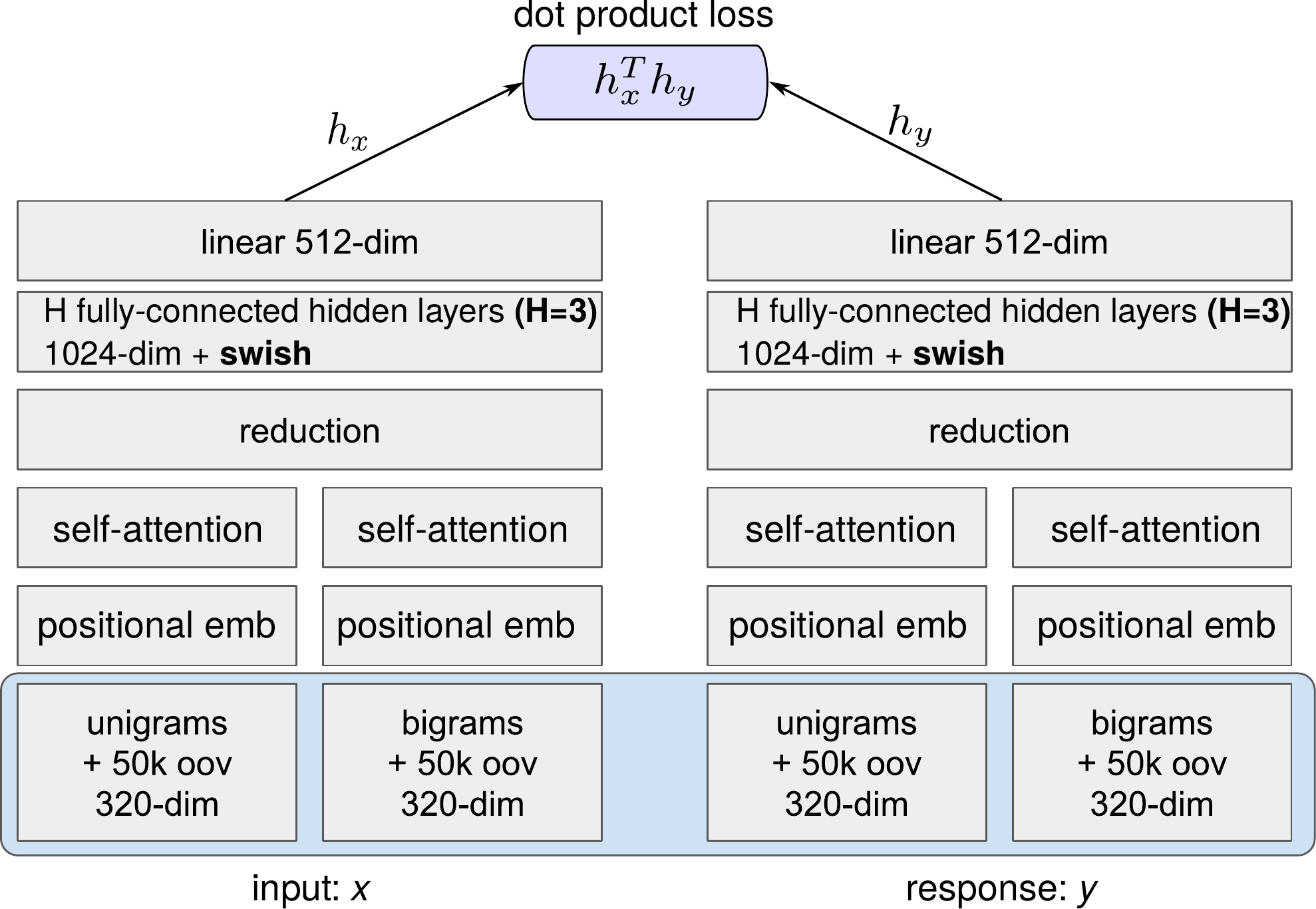}
\vspace{-2.5mm}
\caption{Schematic input-response encoder model structure. We show the best-performing architecture for brevity, while we evaluate a variety of other encoder architecture configurations later in \S\ref{ss:res-reddit}.}
\vspace{-10.5mm}
\label{fig:pretrain}
\end{figure}

\paragraph{Input and Response Representation.}
We now turn to describing the architecture of the main pretraining model. The actual description focuses on the best-performing architecture shown in Figure~\ref{fig:pretrain}, but we also provide a comparative analysis of other architectural choices later in \S\ref{ss:res-reddit}. 

First, similar to \newcite{Henderson:2017arxiv}, raw text is converted to unigrams and bigrams, that is, we extract $n$-gram features from each input $x$ and its corresponding response $y$ from (Reddit) training data. During training we obtain $d$-dimensional feature representations ($d=320$, see Figure~\ref{fig:pretrain}) shared between inputs and responses for each unigram and bigram jointly with other neural net parameters. In addition, the model can deal with out-of-vocabulary unigrams and bigrams by assigning a random id from 0 to 50,000 to each, which is then used to look up their embedding. When fine-tuning, this allows the model to learn representations of words that otherwise would be out-of-vocabulary. 

\paragraph{Sentence Encoders.}
The unigram and bigram embeddings then undergo a series of transformations on both the input and the response side, see Figure~\ref{fig:pretrain} again. Following the transformer architecture \cite{Vaswani:2017nips}, positional embeddings and self-attention are applied to unigrams and bigrams separately. The representations are then combined as follows (i.e., this refers to the \textit{reduction} layer in Figure~\ref{fig:pretrain}): the unigram and bigram embeddings are each summed and divided by the square root of the word sequence length. The two vectors are then averaged to give a single $320$-dimensional representation of the text (input or response).

The averaged vector is then passed through a series of $H$ fully connected $h$-dim feed-forward hidden layers ($H=3$; $h=1,024$) with \textit{swish} as the non-linear activation, defined as: $swish(x)=x \cdot \text{sigmoid}(\beta x)$ \cite{Ramachandran:2017arxiv}.\footnote{We fix $\beta=1$ as suggested by \newcite{Ramachandran:2017arxiv}. The use of \textit{swish} is strictly empirically driven: it yielded slightly better results in our preliminary experiments than the alternatives such as \textit{tanh} or a family of LU/ReLU-related activations \cite{He:2015iccv,Klambauer:2017arxiv}.} The final layer is linear and maps the text into the final $l$-dimensional ($l=512$) representation: $h_x$ for the input text, and $h_y$ for the accompanying response text. This provides a fast encoding of the text, with some sequential information preserved.\footnote{Experiments with higher-order n-grams, recurrent, and convolutional structures have not provided any substantial gain, and slow down the encoder model considerably.}
\begin{figure*}[t]
    \centering
    \begin{subfigure}[t]{0.275\linewidth}
        \centering
        \includegraphics[width=0.98\linewidth]{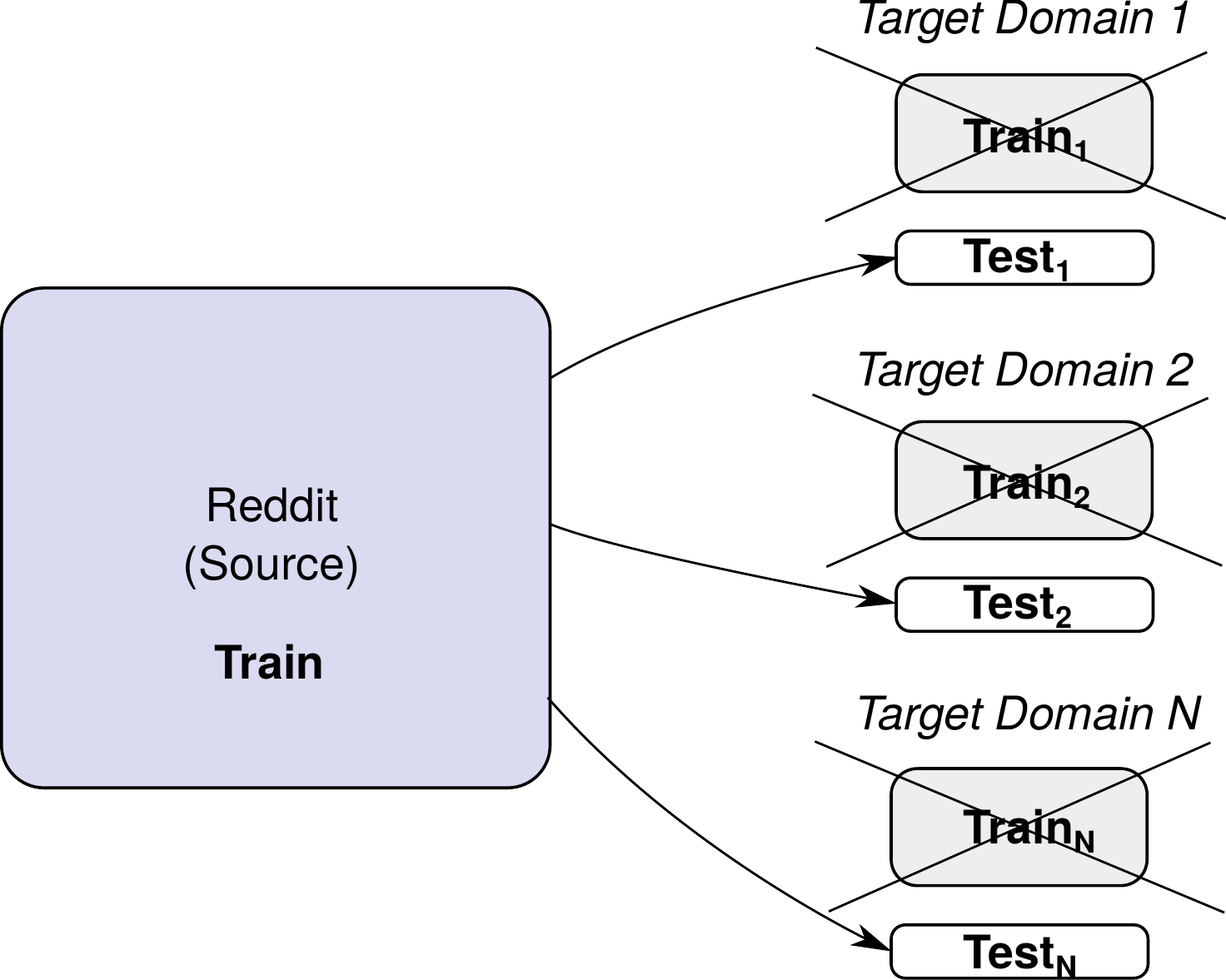}
        \caption{\textsc{reddit-direct}}
        \label{fig:source}
    \end{subfigure}%
    \hspace{1em}
    \begin{subfigure}[t]{0.275\textwidth}
        \centering
        \includegraphics[width=0.98\linewidth]{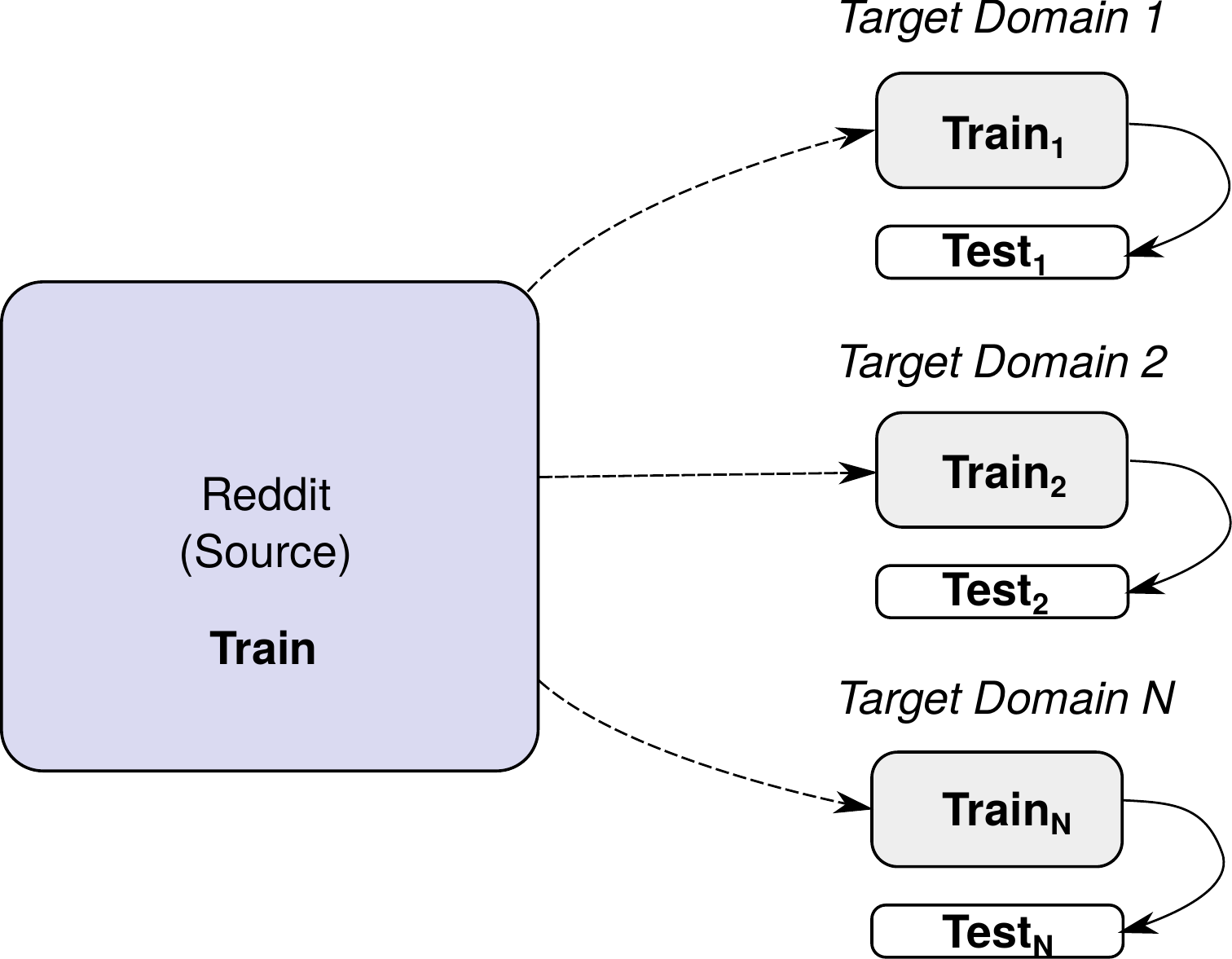}
        \caption{\textsc{ft-direct}}
        \label{fig:direct}
    \end{subfigure}%
    \hspace{1em}
        \begin{subfigure}[t]{0.35\textwidth}
        \centering
        \includegraphics[width=0.96\linewidth]{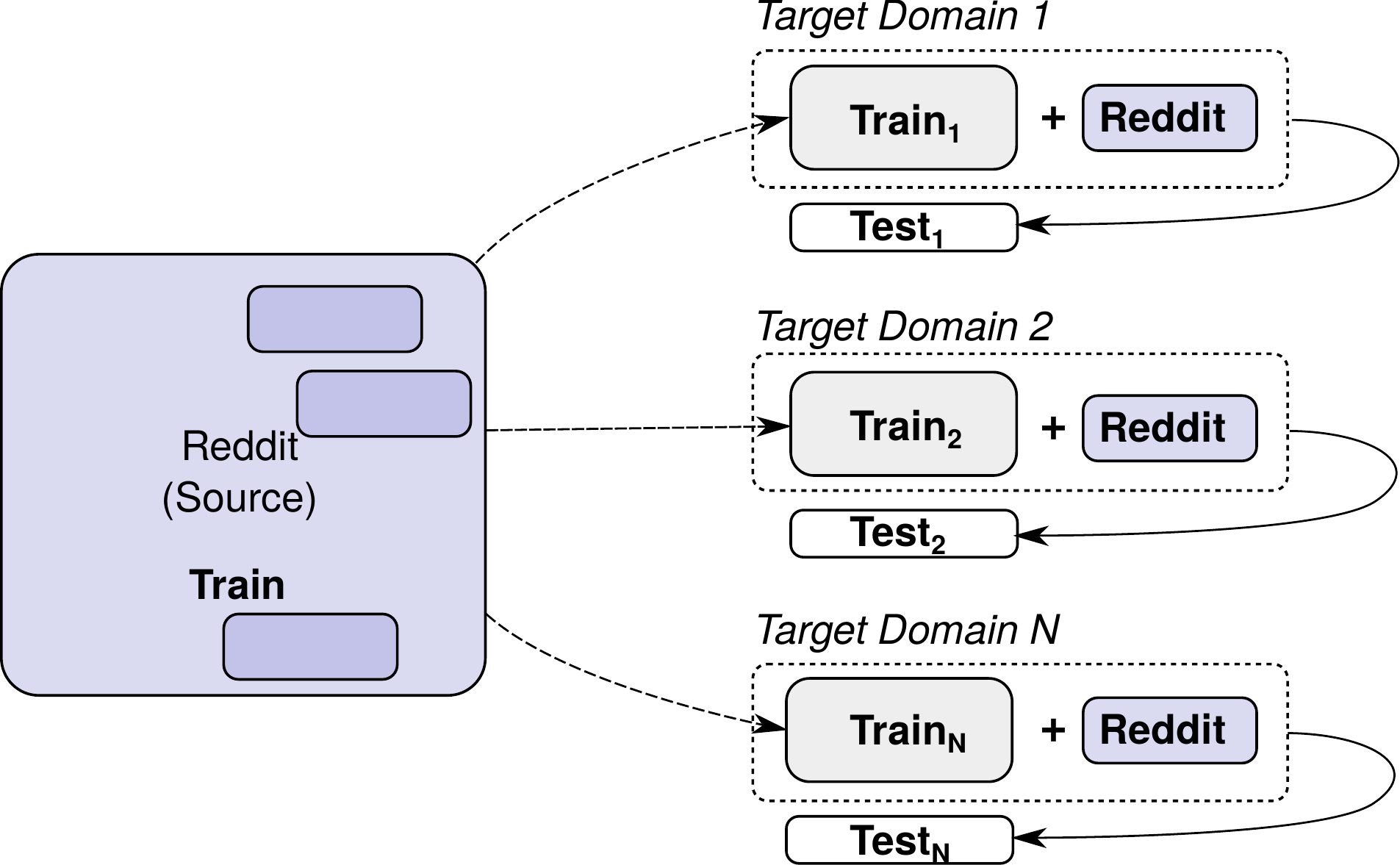}
        \caption{\textsc{ft-mixed}}
        \label{fig:mixed}
    \end{subfigure}
    \vspace{-0.0em}
	\caption{High-level overview of baseline and fine-tuning strategies used in our evaluation. \textbf{(a)} \textsc{reddit-direct}: a pretrained general-domain (Reddit) response selection model is directly applied on each target task, without any target domain fine-tuning; \textbf{(b)} \textsc{ft-direct}: after pretraining the large response selection model on Reddit, the model is fine-tuned for each target task by directly continuing the training on (much smaller) target domain data; \textbf{(c)} \textsc{ft-mixed}: similar to \textsc{ft-direct}, but the crucial difference is in-batch mixing of Reddit input-response pairs with target domain pairs during the target fine-tuning procedure. Another baseline (\textsc{target-only}) trains a response selection model on each target task separately without leveraging general-domain Reddit data (not shown).}
\vspace{-0.5mm}
\label{fig:main}
\end{figure*}

\paragraph{Scaled Cosine Similarity Scoring.} The relevance of each response to the given input is then quantified by the score $S(x,y)$. It is computed as scaled cosine similarity: $S(x,y)=C \cdot cos(h_x,h_y)$, where $C$ is a learned constant, constrained to lie between $0$ and $\sqrt{l}$. We resort to scaled cosine similarity instead of general dot product as the absolute values are meaningful for the former. In consequence, the scores can be thresholded, and retrained models can rely on the same thresholding.

Training proceeds in batches of $K$ \textit{(input, response)} pairs $(x_1,y_1), \ldots, (x_K,y_K)$. The objective tries to distinguish between the true relevant response and irrelevant/random responses for each input sentence $x_i$. The training objective for a single batch of $K$ pairs is as follows:

\vspace{-1.5mm}
\begin{align}
J = \sum_{i=1}^K S(x_i,y_i) - \sum_{i=1}^K \log \sum_{j=1}^{K} e^{S(x_i,y_j)}
\label{eq:training}
\end{align}
%
%
Effectively, Eq.~\eqref{eq:training} maximises the score of pairs $(x_i, y_i)$ that go together
in training, while minimising the score of pairing each input $x_i$ with $K'$ negative examples, that is, responses that are not associated with the input $x_i$. For simplicity, as in prior work \cite{Henderson:2017arxiv,Yang:2018repl}, for each input $x_i$, we treat all other $K-1$ responses in the current batch $y_j \neq y_i$ as negative examples.\footnote{Note that the matrix $\mathbf{S}=C \cdot [h_{y_1}, \ldots, h_{y_K}] \cdot [h_{x_1}, \ldots, h_{x_K}]^T$ is inexpensive to compute.}


As discussed by \newcite{Henderson:2017arxiv} in the context of e-mail reply applications, this design enables efficient response search as it allows for precomputing vectors of candidate responses independently of input queries, and searching for responses with high scaled cosine similarity scores in the precomputed set. It also allows for approximate nearest neighbour search \cite{Malkov:2016arxiv} which speeds up computations drastically at the modest decrease of retrieval performance.\footnote{E.g., experiments on Reddit test data reveal a $130\times$ speed-up using the approximate search method of \newcite{Malkov:2016arxiv} while retaining 95\% top-30 recall.}

Finally, in this work we rely on a simple strategy based on random negative examples. In future work, we plan to experiment with alternative (non-random) negative sampling strategies. For instance, inspired by prior work on semantic specialisation \cite{Mrksic:17} and parallel corpora mining \cite{Guo:2018wmt}, difficult negative examples might comprise invalid responses that are semantically related to the correct response (measured by e.g. dot-product similarity).

\subsection{Step 2: Target Domain Fine-Tuning}
\label{ss:fine-tuning}
The second step concerns the application of the pretrained general Reddit model on $N$ target domains. We assume that we have the respective training and test sets of $K_{N,tr}$ and $K_{N,te}$ in-domain input-response pairs for each of the $N$ domains, where $K_{N,tr}$ and $K_{N,te}$ are considerably smaller than the number of Reddit training pairs. We test two general fine-tuning strategies, illustrated in Figure~\ref{fig:main}.

\textsc{ft-direct} directly continues where the Reddit pretraining stopped: it fine-tunes the model parameters by feeding the $K_{N,tr}$ in-domain \textit{(input, response)} pairs into the model and by following exactly the same training principle as described in \S\ref{ss:pretraining}. The fine-tuned model is then tested in the in-domain response selection task using $K_{N,te}$ test pairs, see Figure~\ref{fig:direct}.

\textsc{ft-mixed} attempts to prevent the ``specialisation'' of the Reddit model to a single target domain, that is, it aims to maintain stable performance on the general-domain Reddit data. This way, the model can support multiple target tasks simultaneously. Instead of relying only on in-domain training pairs, we now perform in-batch mixing of Reddit pairs with in-domain pairs: $M$\% of the pairs in each batch during fine-tuning are Reddit pairs, while $(100-M)$\% of the pairs are in-domain pairs, where $M$ is a tunable hyper-parameter. With this fine-tuning strategy, outlined in Figure~\ref{fig:mixed}, each dataset provides negative examples for the other one, enriching the learning signal.

We compare \textsc{ft-direct} and \textsc{ft-mixed} against two straightforward and insightful baselines: the \textsc{reddit-direct} model from Figure~\ref{fig:source} directly applies the pretrained Reddit model on the target task without any in-domain fine-tuning. Comparisons to this baseline reveal the importance of fine-tuning. On the other hand, the \textsc{target-only} baseline simply trains the response selection model from Figure~\ref{fig:pretrain} from scratch directly on the in-domain $K_{N,tr}$ pairs. Comparisons to this baseline reveal the importance of Reddit pretraining. For all \textsc{target-only} models in all target tasks, we tuned the word embedding sizes and embedding dropout rates on the corresponding training sets.

\section{Experimental Setup}
\label{s:experimental}
\paragraph{Training Setup and Hyper-Parameters.}
All input text is lower-cased and tokenised, numbers with 5 or more digits get their digits replaced by a wildcard symbol \textit{\#}, while words longer than 16 characters are replaced by a wildcard token LONGWORD. Sentence boundary tokens \textit{<S>} and \textit{</S>} are added to each sentence. The vocabulary consists of the unigrams that occur at least 10 times in a random 1M subset of the Reddit training set --this results in a total of 105K unigrams-- plus the 200K most frequent bigrams in the same random subset.

The following training setup refers to the final Reddit model, illustrated in Figure~\ref{fig:pretrain}, and used in fine-tuning. The model is trained by SGD setting the initial learning rate to 0.03, and then decaying the learning rate by 0.3x every 1M training steps after the first 2.5M steps. Similar to learning rate scaling by the batch size used in prior work \cite{Goyal:2017arxiv,Codreanu:2017arxiv}, we scale the unigram and bigram embedding gradients by the batch size. The batch size is 500, and attention projection dimensionality is 64.


We also apply the label smoothing technique \cite{Szegedy:2016cvpr}, shown to reduce overfitting by preventing a network to assign full probability to the correct training example \cite{Pereyra:2017arxiv}. Effectively, this reshapes Eq.~\eqref{eq:training}: each positive training example in each batch gets assigned the probability of 0.8, while the remaining probability mass gets evenly redistributed across in-batch negative examples. Finally, we train the model on 13 GPU nodes with one Tesla K80 each for 18 hours: the model sees around 2B examples and it is sufficient for the model to reach convergence.\footnote{Training is relatively cheap compared to other large models: e.g., BERT models \cite{Devlin:2018arxiv} were pre-trained for 4 days using 4 Cloud TPUs (BERT-SMALL) or 16 Cloud TPUs (BERT-LARGE).} Fine-tuning is run by relying on early stopping on in-domain validation data. The ratio of Reddit and in-domain pairs with \textsc{ft-mixed} is set to 3:1 (in favour of Reddit) in all experimental runs. 

\begin{table*}[t]
\centering
\def\arraystretch{0.97}
\vspace{-0.0em}
{\footnotesize
\begin{tabularx}{\linewidth}{l ll XX}
\toprule
{\bf Dataset} & {\bf Reference} & {\bf Domain} &  {Training Size} & {Test Size} \\
\cmidrule(lr){2-5}
{\textsc{reddit}} & {\cite{Henderson:2019arxiv}} & {discussions on various topics} & {654,396,778} & {72,616,937} \\
{\textsc{OpenSub}} & {\cite{Lison:2016lrec}} & {movies, TV shows} & {283,651,561} & {33,240,156} \\
{\textsc{amazonQA}} & {\cite{Wan:2016icdm}} & {e-commerce, retail} & {3,316,905} & {373,007} \\
{\textsc{ubuntu}} & {\cite{Lowe:2017dd}} & {computers, technical chats} & {3,954,134} & {72,763} \\
{\textsc{banking}} & {New} & {e-banking applications, banking FAQ} & {10,395} & {1,485} \\
{\textsc{semeval15}} & {\cite{Nakov:2015semeval}} & {lifestyle, tourist and residential info} & {9,680} & {1,158} \\
\bottomrule
\end{tabularx}
}
\vspace{-1.5mm}
\caption{Summary of all target domains and data. Data sizes: a total number of unique \textit{(input, response)} pairs. Note that some datasets contain many-to-one pairings (i.e., multiple inputs are followed by the same response; \textsc{banking}) and one-to-many pairings (i.e., one input generates more than one plausible response; \textsc{semeval15}).}
\vspace{-1.5mm}
\label{tab:test_data}
\end{table*}

\paragraph{Test Domains and Datasets.} 
We conduct experiments on six target domains with different properties and varying corpora sizes. The diversity of evaluation probes the robustness of the proposed pretraining and fine-tuning regime. The summary of target domains and the corresponding data is provided in Table~\ref{tab:test_data}. All datasets are in the form of \textit{(input, response)} pairs. For \textsc{ubuntu}\footnote{https://github.com/rkadlec/}, \textsc{semeval15}\footnote{http://alt.qcri.org/semeval2015/task3/}, and \textsc{amazonQA}\footnote{http://jmcauley.ucsd.edu/data/amazon/qa/} we use standard data splits into training, dev, and test portions following the original work \cite{Lowe:2017dd,Nakov:2015semeval,Wan:2016icdm}. For the OpenSubtitles dataset (\textsc{OpenSub}) \cite{Lison:2016lrec}, we rely on the data splits introduced by \newcite{Henderson:2019arxiv}. We evaluate pretrained Reddit models on the \textsc{reddit} held-out data: 50K randomly sampled \textit{(input, response)} pairs are used for testing.

We have also created a new FAQ-style dataset in the e-banking domain which includes question-answer pairs divided into 77 unique categories with well-defined semantics (e.g., ``card activation'', ``closing account'', ``refund request''). Such FAQ information can be found in various e-banking customer support pages, but the answers are highly hierarchical and often difficult to locate. Our goal is to test the fine-tuned encoder's ability to select the relevant answers to the posed question. To this end, for each question we have collected 10 paraphrases that map to the same answer. All unique \textit{(question, answer)} pairs are added to the final dataset, which is then divided into training (70\%), validation (20\%) and test portions (10\%), see Table~\ref{tab:test_data}.

\paragraph{Baseline Models.}
Besides the direct encoder model training on each target domain without pretraining (\textsc{target-only}), we also evaluate two standard IR baselines based on keyword matching: 1) a simple \textsc{tf-idf} query-response scoring \cite{Manning:2008ir}, and 2) Okapi \textsc{bm25} \cite{Robertson:2009}.

Furthermore, we also analyse how pretraining plus fine-tuning for response selection compares to a representative sample of publicly available neural network embedding models which embed inputs and responses into a vector space. We include the following embedding models, all of which are readily available online.\footnote{\url{https://www.tensorflow.org/hub}} 
(1) Universal Sentence Encoder of \newcite{Cer:2018arxiv} is trained using a transformer-style architecture \cite{Vaswani:2017nips} on a variety of web sources such as Wikipedia, web news, discussion forums as well as on the Reddit data. We experiment with the base \textsc{use} model and its larger variant (\textsc{use-large}). (2) We run fixed mean-pooling of \textsc{ELMo} contextualised embeddings \cite{Peters:2018naacl} pretrained on the bidirectional LM task using the LM 1B words benchmark \cite{Chelba:2013arxiv}: \textsc{elmo}. (3) We also compare to two variants of the bidirectional transformer model of \newcite{Devlin:2018arxiv} (\textsc{bert-small} and \textsc{bert-large}).\footnote{Note that the encoder architectures similar to the ones used by \textsc{use} can also be used in the Reddit pretraining phase in lieu of the architecture shown in Figure~\ref{fig:pretrain}. However, the main goal is to establish the importance of target response selection fine-tuning by comparing it to direct application of state-of-the-art pretrained encoders, used to encode both input and responses in the target domain.}

We compare to two model variants for each of the above vector-based baseline models. First, the \textsc{sim} method ranks responses according to their cosine similarity with the context vector: it relies solely on pretrained models without any further fine-tuning or adaptation, that is, it does not use the training set at all. The \textsc{map} variant learns a linear mapping on top of the response vector. The final score of a response with vector $h_y$ for an input with vector $h_x$ is the cosine similarity $cos(\cdot,\cdot)$ of the context vector with the mapped response vector:
\begin{align}
cos\big(h_x, \left( W + \alpha I \right) \cdot h_y\big).
\end{align}
$W, \: \alpha$ are parameters learned on a random sample of 10,000 examples from the training set using the same dot product loss from Eq.~\eqref{eq:training}, and $I$ is the identity matrix. Vectors are $\ell_2$-normalised before being fed to the \textsc{map} method. For all baseline models, learning rate and regularization parameters are tuned using a held-out development set.

The combination of the two model variants with the vector-based models results in a total of 10 baseline methods, as listed in Table~\ref{tab:res:ft1}. 

\paragraph{Evaluation Protocol.}
We rely on a standard IR evaluation measure used in prior work on retrieval-based dialogue \cite{Lowe:2017dd,Zhou:2018acl,Chaudhuri:2018conll}: \textit{Recall@k}. Given a set of $N$ responses to the given input/query, where only one response is relevant, it indicates whether the relevant response occurs in the top $k$ ranked candidate responses. We refer to this evaluation measure as $\mathbf{R}_{N}@k$, and set $N=100; k=1$: $\mathbf{R}_{100}@1$. This effectively means that for each query, we indicate if the correct response is the top ranked response between 100 candidates. The final score is the average across all queries.

\section{Results and Discussion}
\label{s:results}
\begin{table}[t]
\centering
\def\arraystretch{0.99}
\vspace{-0.0em}
{\footnotesize
\begin{tabularx}{\linewidth}{l X}
\toprule
{\textbf{Full Reddit Model}} & {\bf 61.3} \\
\hdashline
- {Wider hidden layers; $h=2,048$}, 24h training & {61.1} \\
- {Narrower hidden layers; $h=750$}, 18h training & {60.8} \\
- {Narrower hidden layers; $h=512$} & {59.8} \\
- {Batch size 50 (before 500)} & {57.4} \\
- {$H=2$ (before $H=3)$} & {56.9} \\
- {\textit{tanh} activation (before \textit{swish})} & {56.1} \\
- no label smoothing & {55.3} \\
- no self-attention & {48.7} \\
- remove bigrams & {35.5} \\
\bottomrule
\end{tabularx}
}
\vspace{-1.5mm}
\caption{The results of different encoder configurations on the Reddit test data ($R_{100}@1$ scores $\times 100\%$). Starting from the full model (top row), each subsequent row shows a configuration with one component removed or edited from the configuration from the previous row.}
\vspace{-11.5mm}
\label{tab:reddit}
\end{table}

\newcommand\ftresult[2]{
    \makebox[\widthof{99.9}][r]{#1} (\makebox[\widthof{99.9}][r]{#2})
}
\newcommand\ftresultt[1]{
    \makebox[\widthof{99.9}][r]{#1}
}
\begin{table*}[!t]
\centering
\def\arraystretch{0.95}
\vspace{-0.0em}
{\footnotesize
\begin{tabularx}{\linewidth}{l XXXXXX}
\toprule
{} & {\textsc{Reddit}} & {\textsc{OpenSub}} &  {\textsc{amazonQA}} & {\textsc{ubuntu}} &  {\textsc{banking}} & {\textsc{semeval15}} \\
\cmidrule(lr){2-2} \cmidrule(lr){3-3}  \cmidrule(lr){4-4} \cmidrule(lr){5-5} \cmidrule(lr){6-6} \cmidrule(lr){7-7}
{\textsc{tf-idf}} &          {26.7} & {10.9} & {51.8} & {27.5} & {27.3} & {38.0}   \\
{\textsc{bm25}} &            {27.6} & {10.9} & {52.3} & {19.6} & {23.4} & {35.5}  \\
\hdashline

\addlinespace[1ex] {\textsc{use-sim}} &         {36.6} & {13.6} & {47.6} & {11.5} & {18.2} & {36.0}  \\
{\textsc{use-map}} &         {40.8} & {15.8} & {54.4} & {17.2} & {79.2} & {45.5}  \\
{\textsc{use-large-sim}} &   {41.4} & {14.9} & {51.3} & {13.6} & {27.3} & {44.0} \\
{\textsc{use-large-map}} &   {47.7} & {18.0} & {61.9} & {18.5} & {81.8} & \textbf{56.5}  \\
{\textsc{elmo-sim}} &        {12.5} & \ftresultt{9.5}  & {16.0} & \ftresultt{3.5}  & \ftresultt{6.5}  & {19.5}   \\
{\textsc{elmo-map}} &        {19.3} & {12.3} & {33.0} & \ftresultt{6.2}  & {87.0} & {34.5}  \\

{\textsc{bert-small-sim}} &        {17.1} & {13.8} & {27.8} & \ftresultt{4.1}  & {13.0} & {13.0}  \\
{\textsc{bert-small-map}} &        {24.5} & {17.5} & {45.8} & \ftresultt{9.0}  & {77.9} & {37.5}  \\

{\textsc{bert-large-sim}} &        {14.8} & {12.2} & {25.9} & \ftresultt{3.6}  & {10.4} & {10.0}  \\
{\textsc{bert-large-map}} &        {24.0} & {16.8} & {44.1} & \ftresultt{8.0}  & {68.8} & {34.5}  \\

\hdashline

\addlinespace[1ex] {\textsc{reddit-direct}}    & \textbf{61.3}             & {19.1}    & {61.4}    & \ftresultt{9.6}     & {27.3} & {46.0}     \\
{\textsc{target-only}}      & - & \ftresult{29.0}{18.2}       & \ftresult{83.3}{11.6}       & \ftresult{6.2}{2.3}        & \ftresult{88.3}{1.2} & \ftresult{7.5}{1.1}       \\
{\textsc{ft-direct}}        & -             & \ftresult{\textbf{30.6}}{40.0} & \ftresult{\textbf{84.2}}{30.8} & \ftresult{\textbf{38.7}}{51.9} & \ftresult{\textbf{94.8}}{55.3} & \ftresult{52.5}{55.2}    \\
{\textsc{ft-mixed}}         & -             & \ftresult{25.5}{60.0}   & \ftresult{77.0}{59.6} & \ftresult{38.1}{59.4} & \ftresult{90.9}{59.8}    & \ftresult{\textbf{56.5}}{59.4} \\
\bottomrule
\end{tabularx}
}
\vspace{-1.5mm}
\caption{
Summary of the results ($R_{100}@1$ scores $\times 100\%$) with fine-tuning on all six target domains. Datasets are ordered left to right based on their size. The scores in the parentheses in the \textsc{target-only}, \textsc{ft-direct} and \textsc{ft-mixed} rows give the performance on the general-domain \textsc{reddit} test data. The scores are computed with de-duplicated inputs for \textsc{semeval15} (i.e., the initial dataset links more responses to the same input), and de-duplicated answers for banking.}
\vspace{-2mm}
\label{tab:res:ft1}
\end{table*}


This section aims to provide answers to the two main questions posed in \S\ref{s:intro}: which encoder architectures are more suitable for pretraining (Q1; \S\ref{ss:res-reddit}), and how to adapt/fine-tune the pretrained model to target tasks (Q2; \S\ref{ss:res-finetuning}).
\subsection{Reddit Pretraining}
\label{ss:res-reddit}
The full encoder model is described in \S\ref{ss:pretraining} and visualised in Figure~\ref{fig:pretrain}. In what follows, we also analyse performance of other encoder configurations, which can be seen as ablated or varied versions of the full model. The results on the \textsc{reddit} response selection task are summarised in Table~\ref{tab:reddit}.

\paragraph{Results and Discussion.}
The scores suggest that the final model gets contribution from its multiple components: e.g., replacing \textit{tanh} with the recently proposed \textit{swish} activation \cite{Ramachandran:2017arxiv} is useful, and label smoothing also helps. Despite contradictory findings from prior work related to the batch size (e.g., compare \cite{Smith:2018icl} and \cite{Masters:2018arxiv}), we obtain better results with larger batches. This is intuitive given the model design: increasing the batch size in fact means learning from a larger number of negative examples. The results also suggest that the model saturates when provided with a sufficient number of parameters, as wider hidden layers and longer training times did not yield any substantial gains.

The scores also show the benefits of self-attention and positional embeddings instead of deep feed-forward averaging of the input unigram and bigram embeddings \cite{Iyyer:2015acl}. This is in line with prior work on sentence encoders \cite{Cer:2018arxiv,Yang:2018repl}, which reports similar gains on several classification tasks. Finally, we observe a large gap with the unigram-only model variant, confirming the importance of implicitly representing underlying sequences with $n$-grams \cite{Henderson:2017arxiv,Mrksic:17a}. Following the results, we fix the pretraining model in all follow-up experiments (top row in Table~\ref{tab:reddit}).

\begin{figure*}[t]
    \centering
    \begin{subfigure}[t]{0.48\linewidth}
        \centering
        \includegraphics[width=1.0\linewidth]{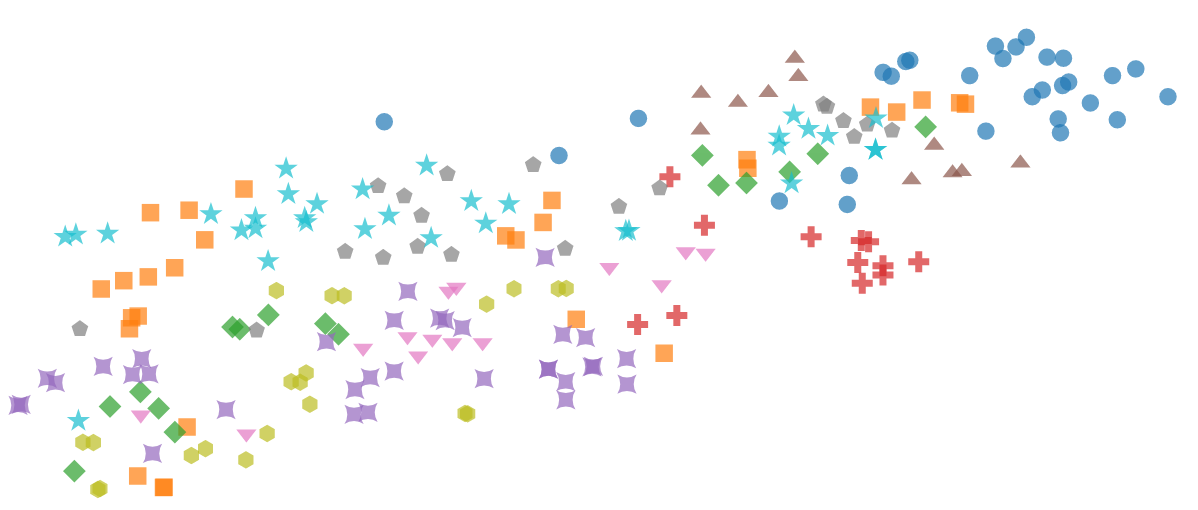}
        \caption{\textsc{elmo-sim}}
        \label{fig:elmo}
    \end{subfigure}%
    \hspace{1em}
    \begin{subfigure}[t]{0.48\textwidth}
        \centering
        \includegraphics[width=1.00\linewidth]{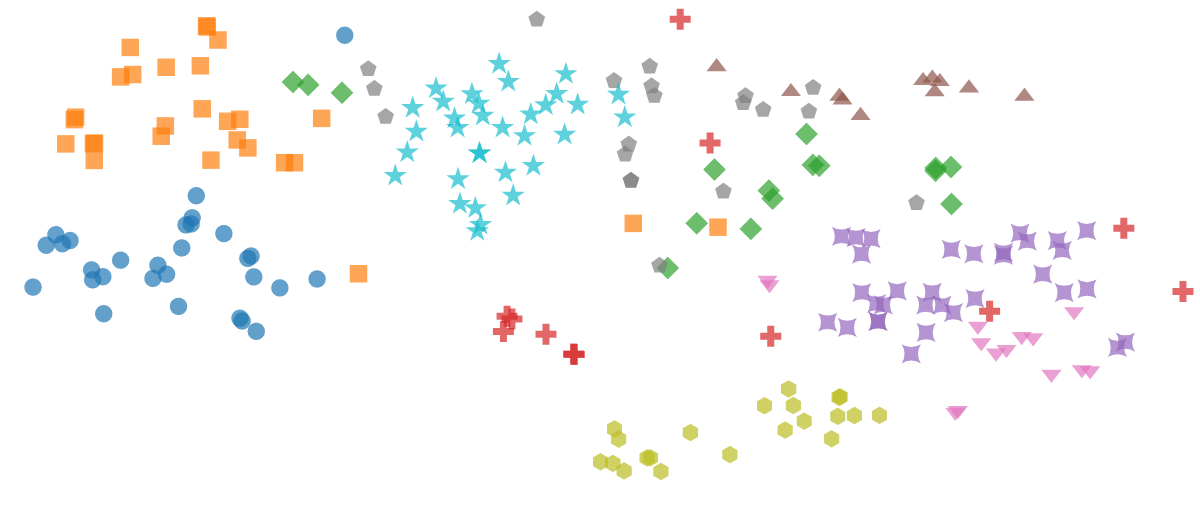}
        \caption{\textsc{use-map}}
        \label{fig:use}
    \end{subfigure}
    \begin{subfigure}[t]{0.48\textwidth}
        \centering
        \includegraphics[width=1.00\linewidth]{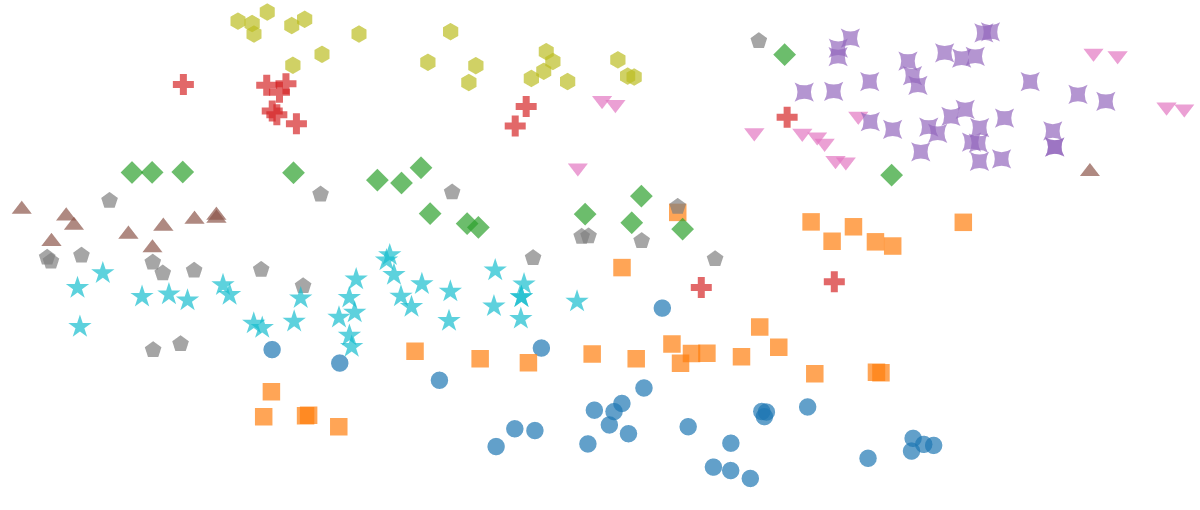}
        \caption{\textsc{reddit-direct} (no fine-tuning)}
        \label{fig:nonfinetuned}
    \end{subfigure}%
    \hspace{1em}
        \begin{subfigure}[t]{0.48\textwidth}
        \centering
        \includegraphics[width=1.0\linewidth]{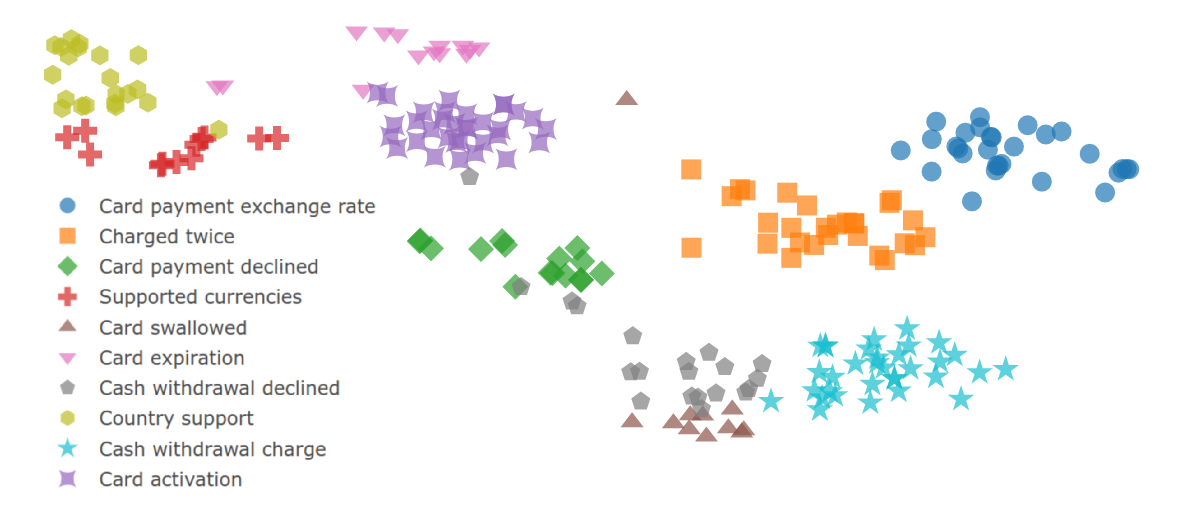}
        \caption{\textsc{ft-mixed} (with fine-tuning)}
        \label{fig:finetuned}
    \end{subfigure}
    \vspace{-0.0em}
	\caption{t-SNE plots \cite{tsne:2012} of encoded questions/inputs for a selection of 10 categories from the \textsc{banking} test set. The most coherent clusters for each category with well-defined semantics are observed with the \textsc{ft-mixed} fine-tuning model applied on top of Reddit response selection pretraining.}
\vspace{-2.5mm}
\label{fig:tsne}
\end{figure*}

\subsection{Target-Domain Fine-Tuning}
\label{ss:res-finetuning}
\noindent \textbf{Results and Discussion.}
The main results on all target tasks after fine-tuning are summarised in Table~\ref{tab:res:ft1}. First, the benefits of Reddit pretraining and fine-tuning are observed in all tasks regardless of the in-domain data size. We report large gains over the \textsc{target-only} model (which trains a domain-specific response selection encoder from scratch) especially for tasks with smaller training datasets (e.g., \textsc{banking}, \textsc{semeval15}). The low scores of \textsc{target-only} with smaller training data suggest overfitting: the encoder architecture cannot see enough training examples to learn to generalise. The gains are also present even when \textsc{target-only} gets to see much more in-domain input-response training data: e.g., we see slight improvements on \textsc{OpenSub} and \textsc{amazonQA}, and large gains on \textsc{ubuntu} when relying on the \textsc{ft-direct} fine-tuning variant.

What is more, a comparison to \textsc{reddit-direct} further suggests that fine-tuning even with a small amount of in-domain data can lead to large improvements: e.g., the gains over \textsc{reddit-direct} are +67.5\% on \textsc{banking}, +32.5\% on \textsc{ubuntu}, +22.8\% on \textsc{amazonQA}, and +11.5\% on \textsc{OpenSub}. These results lead to the following crucial conclusion: while in-domain data are insufficient to train response selection models from scratch for many target domains, such data are invaluable for adapting a pretrained general-domain model to the target domain. In other words, the results indicate that the synergy between the abundant response selection Reddit data and scarce in-domain data is effectively achieved through the proposed training regime, and both components are crucial for the final improved performance in each target domain. In simple words, this finding confirms the importance of fine-tuning for the response selection task.



\paragraph{Comparison to Baselines.}
The results of \textsc{tf-idf} and \textsc{bm25} reveal that lexical evidence from the preceding input can partially help in the response selection task and it achieves reasonable performance across the target tasks. For instance, on some tasks (e.g., \textsc{amazonQA}, \textsc{banking}), such keyword matching baselines even outperform some of the vector-based baseline models, and are comparable to the \textsc{reddit-direct} model variant. They are particularly strong for \textsc{amazonQA} and \textsc{ubuntu}, possibly because rare and technical words (e.g., the product name) are very informative in these domains. However, these baselines are substantially outperformed by the proposed fine-tuning approach across the board. 

A comparison to other pretrained sentence encoders in Table~\ref{tab:res:ft1} further stresses the importance of training for the response selection task in particular. Using off-the-shelf sentence encoders such as \textsc{use} or \textsc{bert} directly on in-domain sentences without distinguishing the input and the response space leads to degraded performance compared even to \textsc{tf-idf}, or the \textsc{reddit-direct} baseline without in-domain fine-tuning. The importance of learning the mapping from input to response versus simply relying on similarity is also exemplified by the comparison between the \textsc{map} method and the simple \textsc{sim} method: regardless of the actual absolute performance, \textsc{map} leads to substantial gains over \textsc{sim} for all vector-based baseline models. However, even the \textsc{map} method cannot match the performance of our two-step training regime: we report substantial gains with our \textsc{ft-direct} and \textsc{ft-mixed} fine-tuning on top of Reddit pretraining for all target domains but one (\textsc{semeval15}).


\paragraph{Further Discussion.}
 The comparison of two fine-tuning strategies suggests that the simpler \textsc{ft-direct} fine-tuning has an edge over \textsc{ft-mixed}, and it seems that the gap between \textsc{ft-direct} and \textsc{ft-mixed} is larger on bigger datasets. However, as expected, \textsc{ft-direct} adapts to the target task more aggressively: this leads to its degraded performance on the general-domain Reddit response selection task, see the scores in parentheses in Table~\ref{tab:res:ft1}. With more in-domain training data \textsc{ft-direct} becomes worse on the \textsc{reddit} test set. On the other hand, \textsc{ft-mixed} manages to maintain its high performance on \textsc{reddit} due to the in-batch mixing used in the fine-tuning process.\footnote{Varying the parameter $M$ in \textsc{ft-mixed} from the ratio 3:1 to 1:3 leads only to slight variations in the final results.}

\paragraph{Qualitative Analysis.}
The effect of fine-tuning is also exemplified by t-SNE plots for the \textsc{banking} domain shown in Figure~\ref{fig:tsne}.\footnote{For clarity, we show the plots with 10 (out of 77) selected categories, while the full plots with all 77 categories are available in the supplemental material.} Recall that in our \textsc{banking} FAQ dataset several questions map to the same response, and ideally such questions should be clustered together in the semantic space. While we do not see such patterns at all with \textsc{ELMo}-encoded questions without mapping (\textsc{elmo-sim}, Figure~\ref{fig:elmo}), such clusters can already be noticed with \textsc{use-map} (Figure~\ref{fig:use}) and with the model pretrained on Reddit without fine-tuning (Figure~\ref{fig:nonfinetuned}). However, fine-tuning yields the most coherent clusters by far: it attracts encodings of all similar questions related to the same category closer to each other in the semantic space. This is in line with the results reported in Table~\ref{tab:res:ft1}.

\section{Related Work}
\label{s:related}
\textbf{Retrieval-Based Dialogue Systems.}
Retrieval-based systems \cite[\textit{inter alia}]{Yan:2016sigir,Bartl:2017arxiv,Wu:2017acl,Song:2018ijcai,Weston:2018ws} provide less variable output than generative dialogue systems \cite{Wen:2015emnlp,Wen:2017icml,vinyals:15}, but they offer a crucial advantage of producing more informative, semantically relevant, controllable, and grammatically correct responses \cite{Ji:2014arxiv}. Unlike modular and end-to-end task-oriented systems \cite{young:10b,Wen:17,Mrksic:2018acl,Li:2018arxiv}, they do not require expensive curated domain ontologies, and bypass the modelling of complex domain-specific decision-making policy modules \cite{Gasic:2015asru,Chen:2017kdd}. Despite these desirable properties, their potential has not been fully exploited in task-oriented dialogue.

Their fundamental building block is response selection \cite{Banchs:2012acl,Wang:2013emnlp,AlRfou:2016arxiv,Baudis:2016arxiv}. We have witnessed a recent rise of interest in neural architectures for modelling response selection \cite{Wu:2017acl,Chaudhuri:2018conll,Zhou:2018acl,Tao:2019wsdm}, but the progress is still hindered by insufficient domain-specific training data \cite{ElAsri:2017sigdial,Budzianowski:2018emnlp}. While previous work typically focused on a single domain (e.g., Ubuntu technical chats \cite{Lowe:15,Lowe:2017dd}), in this work we show that much larger general-domain Reddit data can be leveraged to pretrain response selection models that support more specialised target dialogue domains.

To the best of our knowledge, the work of \newcite{Henderson:2017arxiv} and \newcite{Yang:2018repl} is closest to our response selection pretraining introduced in \S\ref{ss:pretraining}. However, \newcite{Henderson:2017arxiv} optimise their model for one single task: replying to e-mails with short messages \cite{Kannan:2016kdd}. They use a simpler feed-forward encoder architecture and do not consider wide portability of a single general-domain response selection model to diverse target domains through fine-tuning. \newcite{Yang:2018repl} use Reddit conversational context to simply probe semantic similarity of sentences \cite{Agirre:2012semeval,Agirre:2013semeval,Nakov:2016semeval}, but they also do not investigate response selection fine-tuning across diverse target domains.


\paragraph{Pretraining and Fine-Tuning.}
Task-specific fine-tuning of language models (LMs) pretrained on large unsupervised corpora \cite{Peters:2018naacl,Devlin:2018arxiv,Howard:2018acl,Radford:2018,Lample:2019arxiv,Liu:2019arxiv} has taken NLP by storm. Such LM-based pretrained models support a variety of NLP tasks, ranging from syntactic parsing to natural language inference \cite{Peters:2018naacl,Devlin:2018arxiv}, as well as machine reading comprehension \cite{Nishida:2018cikm,Xu:2019arxiv} and information retrieval tasks \cite{Nogueira:2018arxiv,Yang:2019arxiv}. In this work, instead of the LM-based pretraining, we put focus on the response selection pretraining in particular, and show that such models coupled with target task fine-tuning \cite{Howard:2018acl} lead to improved modelling of conversational data in various domains. 

\vspace{-0.5mm}
\section{Conclusion and Future Work}
\label{s:conclusion}
We have presented a novel method for training neural response selection models for task-oriented dialogue systems. The proposed training procedure overcomes the low-data regime of task-oriented dialogue by pretraining the response selection model using general-domain conversational Reddit data and efficiently adapting this model to individual dialogue domains using in-domain data. Our evaluation demonstrates the compelling benefits of such pretraining, with the proposed training procedure achieving strong performance across each of the five different dialogue domains. In future work, we will port this approach to additional target domains, other languages, and investigate more sophisticated encoder architectures and fine-tuning strategies.

\bibliographystyle{acl_natbib}
\bibliography{acl2019_vr_refs}

\end{document}